# Benchmarking quantized LLaMa-based models on the Brazilian Secondary School Exam

1st Matheus L. O. Santos
*Systems and Computing Department*
*Federal University of Campina Grande*
Campina Grande, Brazil
matheus.lisboa.santos@ccc.ufcg.edu.br

2nd Cláudio E. C. Campelo
*Systems and Computing Department*
*Federal University of Campina Grande*
Campina Grande, Brazil
campelo@dsc.ufcg.edu.br

*Abstract*—Although Large Language Models (LLMs) represent a revolution in the way we interact with computers, allowing the construction of complex questions and the ability to reason over a sequence of statements, their use is restricted due to the need for dedicated hardware for execution. In this study, we evaluate the performance of LLMs based on the 7 and 13 billion LLaMA models, subjected to a quantization process and run on home hardware. The models considered were Alpaca, Koala, and Vicuna. To evaluate the effectiveness of these models, we developed a database containing 1,006 questions from the ENEM (Brazilian National Secondary School Exam). Our analysis revealed that the best performing models achieved an accuracy of approximately 46% for the original texts of the Portuguese questions and 49% on their English translations. In addition, we evaluated the computational efficiency of the models by measuring the time required for execution. On average, the 7 and 13 billion LLMs took approximately 20 and 50 seconds, respectively, to process the queries on a machine equipped with an AMD Ryzen 5 3600x processor.

*Index Terms*—Large language models, LLMs, ENEM, GGML, LLaMA, Quantization.

## I. INTRODUCTION

With the introduction of the article *Attention is all you need* [1], the field of Natural Language Processing (NLP) underwent a significant revolution. Tasks that were previously dominated by heuristics and machine learning algorithms began to achieve state-of-the-art results with the use of Transformers [2]. This neural network architecture aims to pay attention to the most relevant parts of the inputs, such as keywords or areas with people in an image, for example.

With the emergence of *transformers*, a class of neural network models that are trained to predict the next word given a sequence of previous words had their metrics elevated to the state of the art. This category of models is known as language models, and its first applications were aimed at generating *word embeddings* [3]. This technique makes it possible to dynamically assign words to a semantic vector space, where similar words are close to each other. Later, encoder-decoder architectures known as *seq2seq* were used, which made use of *transformers* to achieve state-of-the-art text encoding and decoding tasks. A notable example is the translation of texts between different languages, even when these texts are of different lengths.

With the introduction of the GPT (Generative Pre-trained Transformer) family of models, models trained through unsupervised learning gained popularity. These models were pre-trained on large amounts of unlabeled data and retained general knowledge during their training. They were then fine-tuned on a much smaller amount of data and for shorter periods of time for specific tasks. However, the release of Chat-GPT, a model trained for human interactions through conversations, brought even greater visibility to these models.

These models have brought significant innovation in the way humans interact with computers, enabling intuitive communication through dialogues where the responses are precisely tailored to the requests. This results in significant time savings compared to traditional search on search engines. However, it is important to note that these models are not freely accessible. For example, the renowned Chat-GPT model does not publicly provide its source code, which prevents researchers from conducting studies on its internal workings. Additionally, access to its functionalities through the API requires payment of fees.

However, companies such as Meta[1] have taken an open-source approach by making Large Language Models (LLMs) available as a basis for researchers and enthusiasts to conduct their research. The models released by Meta have sizes of 7, 13, 30, and 65 billion parameters for the first version, and 7, 13 and 70 billion for the second version. Although these models are considered smaller compared to the GPT family (for example, GPT-3.5 Turbo has 154 billion parameters), it still requires dedicated hardware to run them, which restricts research to people who have access to these resources.

However, as has been shown by [4], it is possible to decrease the amount of memory required to use these models with a quantization process. This process aims to decrease the accuracy of the weights of the hidden layers of the models at the cost of performance loss. Using quantization techniques, one project aims to use an API written from scratch in C/C++ for model execution without the need for dedicated GPUs[2]. The models are based on LLaMA, published by Meta [5], they are: Vicuna[3], Koala[4] and Alpaca[5], all of which have two variants, one of 7 and one of 13 billion parameters. This allowed anyone to experience the potential of these models, since it would be possible to run inference on domestic hardware.

---
[1]https://about.meta.com/

The Brazilian National Secondary School Exam (ENEM) is a test that is taken annually by secondary school students across the country and serves as a gateway to colleges throughout Brazil, thus representing a challenge that many students prepare for all year long. As demonstrated by [6], these LLMs are able to generalize knowledge, increasing the number of activities they perform as they increase the number of parameters. That said, evaluating the performance of these LLMs on ENEM questions becomes a good benchmark of how robust these large models are, since these are general purpose models, and have not been trained to answer questions.

Hence, the goal of this study is to evaluate quantized language models, based on LLaMA [5], capable of operating on home hardware, using ENEM questions as the analysis scenario. For this purpose, we produced a carefully structured database of questions containing the texts of the questions along with the correct answers. The database encompasses a total of 1,006 questions, covering the period from 2010 to 2022. The database produced has great potential for LLM analysis and also for other studies in the field of natural language processing.

The experiments conducted in our study aim to answer the following research questions:

- **Q1** - How effective are quantized models, based on LLaMA, trained in English, in solving ENEM questions described in Brazilian Portuguese?
- **Q2** - How effective are quantized models, based on LLaMA, trained in English, in solving ENEM questions translated from Brazilian Portuguese into English?
- **Q3** - There is an improvement between the LLaMA models from the first version to the second?
- **Q4** - How efficient (in terms of time to run on a computer with modest hardware) are quantized models, based on LLaMA, when used to solve ENEM questions?

## II. Related Work

The use of LLMs is rapidly advancing in various research fields. One notable application is in the field of medicine, where researchers utilized the PALM model [7], trained by Google, to perform question answering in the medical domain. This model was evaluated on the United States Medical Licensing Examination (USMLE) [8], and the analysis demonstrated that the model provided answers that reached a consensus with experts in 92.6% of the questions. This highlights the potential benefits of these models in assisting healthcare professionals in their practice.

As shown by [9], there are already efforts in training LLMs for question solving. According to the comparative study provided by the authors, their model performed better than all other models available on the market, except for GPT-4, for English and Chinese exams. The model was evaluated on the following datasets: MMLU, AGIEval, and C-Eval, and had the following metrics: 67.2, 49.2, and 62.7, respectively; Against 86.2, 56.4, and 68.7 from GPT-4.

Additionally, there are reports of research in training language models with a focus on creating a chain of thought, where the model is able to explain the why of its responses [10]. This can help create language models that are increasingly able to provide responses that are useful to humans. Thinking in a question answering context, a model that was able to explain the reasoning behind the answer to an alternative would be very useful to a student, for example.

In the Brazilian context, a team of researchers proposed to use GPT-4 [11] to evaluate questions from ENEM [12]. The model showed 87.29% accuracy on the 2022 questions, against 73.73% accuracy of gpt-3.5-turbo. This improvement was due to the increase in the size of the model to also include images. This shows that these models were able to perform better than most of the humans who take this exam every year.

Quantized language models are in focus, given the number of computational resources required to run them [13], [14]. However, these studies address evaluation using abstract metrics[6]. This work aims to evaluate these quantized models in a tangible way, checking how well they can answer a challenging test such as ENEM.

## III. Theoretical Framework

This section introduces some relevant concepts for a better understanding of the rest of the article.

### A. Large Language Models – LLMs

One of the determining factors for the high efficiency exhibited by some language models is their size [6]. For example, the GPT-3 [15] model, published by OpenAI[7], has 175 billion parameters, resulting from 34 days of training on 1,024 Nvidia A100 GPUs. The estimated cost for this training was $4.6 million.

For comparison, the 7 billion parameters LLaMA [5] model published by Meta requires a GPU with at least 28 GB of memory to run an inference[8]. These requirements are prohibitive, as such equipment is expensive.

### B. LLaMA.cpp

LLaMA.cpp[9] is a project that aims to create an API for CPU inference of LLMs, using C/C++ and techniques that allow models not to be loaded completely into memory. The templates are based on LLaMA [5], and can run on home computers.

---

[2]https://github.com/ggerganov/llama.cpp
[3]https://lmsys.org/blog/2023-03-30-vicuna/
[4]https://bair.berkeley.edu/blog/2023/04/03/koala/
[5]https://crfm.stanford.edu/2023/03/13/alpaca.html
[6]https://huggingface.co/docs/transformers/perplexity
[7]https://openai.com/
[8]https://discuss.huggingface.co/t/llama-7b-gpu-memory-requirement/34323
[9]https://github.com/ggerganov/llama.cpp

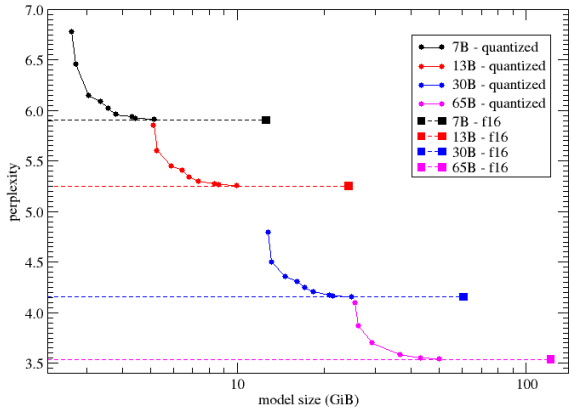

Fig. 1. Performance degradation of quantized models. Chart available at: https://github.com/ggerganov/llama.cpp/pull/1684.

However, it is important to note that these benefits are not obtained without costs. To enable the execution of LLaMA.cpp, it is necessary to reduce the size of the models, which is achieved by applying a quantization technique. This technique involves compressing the weights in the hidden layers of the models, resulting in a reduction in the space required for their storage. Figure 1 illustrates that as the level of quantization increases, i.e., there is a loss of precision in the layers, the perplexity metric increases.

To conduct the experiments described in this paper, all models were quantized at Q4. According to the authors of the repository, this level of quantization leads to a worsening of the perplexity metric by about 2%. More details about the quantization process can be found in Section III-C.

### C. Model quantization process

The process of quantizing the models used in LLaMA.cpp is described in the project Ggml[10]. This project aims to compress different language models, not only those based on LLaMA, by also quantizing models from the GPT family, such as GPT-2 and GPT-J.

The weights of the hidden layers in a model without quantization are represented as 16-bit floats. In the quantization process described in [11], a set of *QK* weights are represented as an integer part plus a floating point part. For example, for a quantization of *Q4*, a block of 4 weights, each being represented in float16, are represented as a float32 scale factor plus 2 integers of 2 bytes each. According to the author, this approach reduced the size of the models by 75%.

[10]https://github.com/ggerganov/ggml
[11]https://github.com/ggerganov/ggml/pull/27

## IV. METHODOLOGY

This section presents the methodology adopted to evaluate the models. It discusses how the database for evaluation was made, the models used, and the experiments conducted.

### A. Dataset

One of the main contributions of this paper is the provision of a structured and validated database composed of numerous questions from the Brazilian Secondary School Exam (ENEM) [16].

The questions basically consist of three parts: the first being a portion of text, tables or images, or a combination of these. The second part is a question about the first part. And finally, five alternatives, only one of which is correct.

This database was developed with a focus on questions that can only be answered by text, since the models that will be evaluated have textual comprehension capabilities. In total, the database contains 1,006 questions, in which the description texts, the alternatives, and the correct answers were identified. The process of collecting these questions followed the following procedure:

- Collection of ENEM tests, from 2010 to 2022, in PDF format, obtained from Instituto Nacional de Estudos e Pesquisas Educacionais Anísio Teixeira (INEP)[12].
- Use of the tool[13] for text extraction from each PDF file.
- Definition of heuristics for concatenating the text of each question, grouping description, question, and alternatives.
- Filtering out questions that did not fit the scope of the experiments.

The following criteria were established for removing questions not suitable for the experiments:

- Questions containing some image, table, or equation; since the models we will use can only understand text.
- Any question that it was not possible to distinguish which parts of the text were the alternatives, since this part was of utmost importance for the models.
- Questions that were not processed properly by the PDF file content extraction tool. These questions had, for example, strange characters in their content.

To remove questions that contain images, tables, or equations, heuristics were used to check if within the question there are any of the keywords, such as: **table, figure, image**. With this, we were able to remove many questions that would be impossible for the models to answer.

The distribution of these questions by year is shown in Figure 2. No questions were extracted for the years 2010 and 2021 due to problems in reading the PDF. The distribution of questions by subject area can be seen in Figure 3. In this figure, it is possible to see that mathematics and natural sciences and their technologies were the areas with the fewest questions due to the filtering of questions that contain graphs, equations, and tables.

[12]https://www.gov.br/inep/pt-br/areas-de-atuacao/avaliacao-e-exames-educacionais/enem/provas-e-gabaritos
[13]https://pymupdf.readthedocs.io/en/latest/document.html

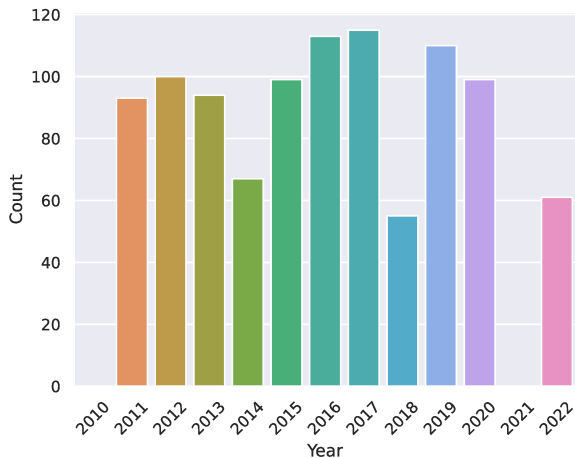

Fig. 2. Count of questions extracted per year.

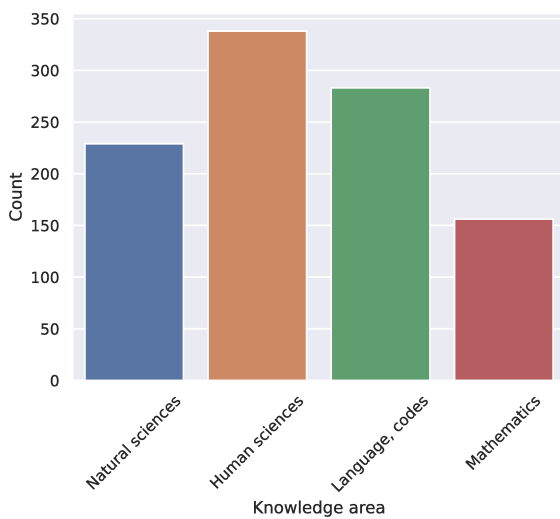

Fig. 3. Distribution of questions extracted per knowledge area.

The annotation of the answers was performed manually, based on the ground truth available in PDF format, being registered in a file in JSON format. We preferred a manual approach, since implementing a script for automation would be costly, since the PDF files have different structures.

The dataset produced is freely available, as well as the artifacts used for its production (files in PDF format and source code of the data processing and transformation scripts) at[14].

### B. Models evaluated

Language models were selected that were aligned with the goal of the study, i.e., large models capable of running on home machines. The models were obtained from the Hugging Face[15] model repository, and were made available by users who performed the quantization process. The models were tested to verify that they are compatible with LLaMA.cpp[16]. This tool provides the execution of models based on LLaMA [5] on domestic machines by employing quantization techniques and selective reading of the parts needed for model execution.

For the experiments, LLaMA v1 and v2 based models of 7 and 13 billion parameters, resulting from fine-tuning of the original models, were used. These are:

- **LLaMA 1 7b, 13b**: Models trained from scratch on a diverse dataset that comes from various sources. These are: **English Common Crawl, C4, GitHub, Wikipedia, Gutenberg and Books3, arXiv and Stack Exchange**. This dataset contains approximately 1.4 trillion tokens, but for the 7 and 13 billion parameter models, a subset of 1 trillion was used.
- **Alpaca 7b, 13b**: Resulting from fine-tuning the LLaMA models with a set of 52,000 question and answer examples, this model was trained to perform better in question and answer scenarios.
- **Koala 7b, 13b**: Fine-tuning of the LLaMA models, but trained 117,000 user iterations with ChatGPT[17]. This model was trained to perform better in dialogs.
- **Vicuna 7b, 13b**: Fine-tuning the LLaMA models, but trained with a set of 70,000 user iterations with ChatGPT, via the ShareGPT data, which are community-sourced conversations with the model.
- **LLaMA 2 7b, 13b**: Second published version of the LLaMA [17]. According to the authors, an optimized version of the autoregressive model was used, with a more robust data treatment and 40% more training data.

One point to note is that the data used to train these models is mostly in English, and no evidence was found that these models had been exposed to data from ENEM questions during their training or validation stages, which could invalidate the results presented in Section V.

### C. Experiment Definitions

As already discussed, these language models can only receive a portion of text on input and return another portion of text on output, so an integral part of the activity is to define the format of the text that will be used to feed them. For the preparation of the prompts, the methodology proposed in the course Prompt Engineering[18] was adopted, made available by OpenAI, the company that published the GPT family models. Although the course is focused on the GPT models, most of the models we used in this experiment are based on data extracted from conversations with the Chat-GPT, so it is expected that the way that these models work is to some degree similar to the methodologies provided in the course. The approach taken was to ask the model to answer the correct alternative and to flag only the letter of the alternative, to facilitate the verification

---

[14]https://github.com/wineone/tcc-matheus-lisboa
[15]https://huggingface.co/models
[16]https://github.com/ggerganov/llama.cpp
[17]https://openai.com/blog/chatgpt
[18]https://www.deeplearning.ai/short-courses/chatgpt-prompt-engineering-for-developers/

```
What is the correct alternative to the question between <>? Answer onl
y with the letter that represents the alternative:

<Question 98) Spiders, scorpions, ticks and mites are representatives
of the arachnid class. These animals are terrestrial mostly and occupy
the most varied habitats, such as high mountains, marshes, deserts and
sandy soils. They may have been the first representatives of the Arthr
opoda phylum to inhabit the dry land.
The characteristic that justifies the adaptive success of this group i
n the occupation of the terrestrial environment is the presence of
A) Chelices and pedipalpos that coordinate body movement.
B) Excretion of uric acid that gives stability to body pH.
C) exoskeleton consisting of chitin that assists in body water contro
l.
D) open blood circulation that prevents dehydration of bodily tissues.
E) ganglion nervous system that promotes the central coordination of b
ody movement.>
```

Fig. 4. Example of a question that will be used for an inference in the models.

of the effectiveness of the models and the computation of the evaluation metrics. Figure 4 shows an example of a prompt.

To perform the comparison of the models, two experiments were run. The first experiment, aiming to answer the question **Q1**, compared the accuracy of the models by running all the models in all the questions, replacing the text of the questions in the prompt, and collecting the result of the models in the text. The second experiment was designed to answer **Q2**, for this all the questions were translated, as well as the prompt, and all the answers were computed. The Google Translate API was used for the translation, using the Text Blob[19] library.

The execution times for these models were also evaluated in order to answer **Q4**. The evaluation was conducted using two machines, one equipped with an AMD Ryzen 5 3600x processor and the other with an Intel i9 9900k processor. The time in seconds was collected for the inference of the questions in Portuguese and English, with the Portuguese questions executed on the machine equipped with the 3600x and the English questions on the machine equipped with the 9900k. The results are presented in Section V.

### D. Model evaluation

In order to evaluate the assertiveness of the models in answering the test questions, we adopted the metric of accuracy, which is defined as the number of correct questions divided by the total number of questions, as described in Equation 1.

$$acc = \frac{\#correct}{\#total} \quad (1)$$

Model accuracy calculation

One of the problems encountered was how to identify which alternative was predicted by the model, given the generational nature of the text. For the vast majority of the prompts, the model presented a very objective output, containing only one letter representing some possible alternatives (A, B, C, D, E). However, in other situations, the model output consisted of a text about the question, followed by the letter representing the answer. In addition to these, we also observed outputs containing long texts without much meaning and without an objective answer. With this in mind, a set of heuristics was defined to capture the alternative selected by the model. The aim of these heuristics is to identify the alternative predicted by the model from among the text returned by the model. For example, in the text "The answer is B" or "B)" the alternative chosen by the model was B. Table I presents the percentage of questions that we were able to identify as an alternative signaled by the model. A manual inspection was performed to ensure that the heuristics identified all available alternatives.

[19]https://pypi.org/project/textblob/

TABLE I
COVERAGE OF QUESTIONS WITH IDENTIFIED ALTERNATIVE.

| Experiment | % of coverage |
|---|---|
| Execution in Portuguese | 0.993 |
| Execution in English | 0.990 |

### V. RESULTS AND DISCUSSIONS

This section presents and discusses the results observed from the experiments conducted. The research questions will be answered:

- **Q1** - How effective are the models on questions in Portuguese?
- **Q2** - How effective are the models on questions translated into English?
- **Q3** - How was the improvement between LLaMA 1 and LLaMA 2?
- **Q4** - How long does it take to run these models on home machines?

*A. Q1 e Q2 – How effective are the models on questions described in Portuguese and English?*

Addressing the **Q1** and **Q2**, the accuracy of the models on the question set was evaluated. In Table II, the performance of the models is presented. It can be seen that some models, such as LLaMA 1 7b and 13b, Alpaca 7b, Koala 7b and 13b, and LLaMA 2 7b performed similarly to a random classifier. This suggests that these models may not be able to adequately understand the questions and provide the correct answers in both English and Portuguese. However, they demonstrated an ability to recognize that the text provided is a question and were able to indicate an alternative, even if incorrect.

During the inference phase, a bias phenomenon was observed in the models analyzed. Most of these models showed a consistent tendency to generate a single option as a result. The percentage distribution of the questions identified in Portuguese during this phase for each model is illustrated in Figure 5, while the distribution for the English language is represented in Figure 6. Except for the LLaMA 1 7b, Vicuna 7b and Vicuna 13b, LLaMA 1 7b, and LLaMA 2 7b and 13b models, all the others showed a significant bias towards alternative A, contrary to the expectation of a balanced distribution among all options. Notably, the Vicuna 13b model exhibited a bias toward alternative B for both languages, while the LLaMA 1 7b and LLaMA 2 7b models showed a bias toward alternative D in Portuguese, and toward alternatives B

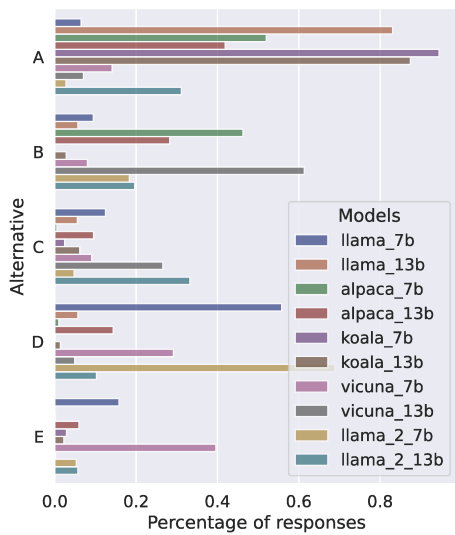

Fig. 5. Distribution of alternatives identified in the models, questions in Portuguese.

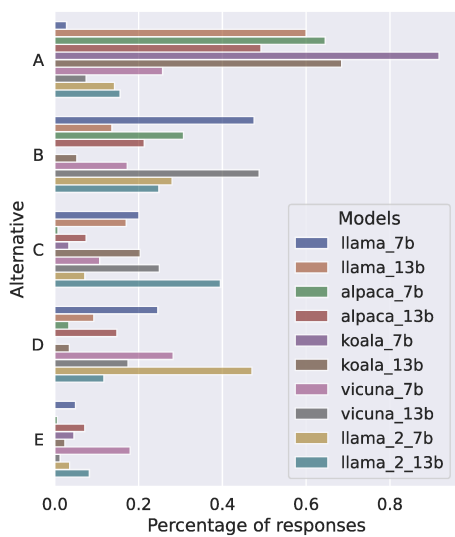

Fig. 6. Distribution of alternatives identified in the models, questions in English.

TABLE II
OVERALL ACCURACY OF THE MODELS.

| Model | Language | |
|---|---|---|
| | Portuguese | English |
| **LLaMA 1 7b** | 0.225 | 0.251 |
| **LLaMA 1 13b** | 0.207 | 0.230 |
| **Alpaca 7b** | 0.203 | 0.205 |
| **Alpaca 13b** | 0.400 | 0.339 |
| **Koala 7B** | 0.183 | 0.193 |
| **Koala 13b** | 0.243 | 0.289 |
| **Vicuna 7b** | 0.327 | 0.399 |
| **Vicuna 13b** | 0.336 | 0.397 |
| **LLaMA 2 7b** | 0.292 | 0.348 |
| **LLaMA 2 13b** | **0.468** | **0.493** |

and D in English, respectively. The 7 billion parameter Vicuna, and the LLaMA 2 13b were identified as the models with the lowest bias, as it did not show a significant bias toward any of the options or languages. Still, the models seemed to show a more pronounced bias toward Portuguese, while they showed a less pronounced bias toward English

However, the Alpaca 13b, Vicuna 7b and 13b, and LLaMA 2 13b models performed significantly better, with the 7b Vicuna achieving an accuracy rate of approximately 40% for the English language and the 13b Alpaca achieving 40% accuracy for the Portuguese language. The best model evaluated was LLaMA 2 with 13 billion parameters, which achieved an accuracy of 46.8% for Portuguese and 49.3% for English. While these results are distant from those reported by [12] for Chat-GPT, they are quite promising, considering that these models are open-source alternatives, have undergone quantization, and can be run on domestic machines without the need for specialized hardware.

As for the assumption that the models would perform better at English than in Portuguese, this was true for LLaMA 1 13b, Koala 7b and 13b, Vicuna 7b and 13b LLaMA 2 7b and 13b. The best metrics for each language were 46.8% for Portuguese, against 49.3% in English, suggesting that there is indeed an improvement in translating the questions and evaluating the models.

To better observe the capacity of the models, the metrics were also compared in the four areas of knowledge of the ENEM test, which are: 'Humanities and its technologies'; 'Nature sciences and its technologies'; 'Mathematics and its technologies'; 'Languages, codes and its technologies'. The metrics can be found in Table III. Both for Portuguese and English, the models managed to perform well in the areas of 'humanities and its technologies' and 'codes and its technologies', with the LLaMA 13b having an accuracy of 63.6% and 51.5%, respectively. In the area of 'natural sciences', the result was a little worse, with the LLaMA 2 13b achieving an accuracy of 41.4%. In the area of 'mathematics and its technologies', no LLaMA model performed satisfactorily, having their accuracies limited to 24.3% for Portuguese and 26.2% for English. Moreover, in this area, the observed accuracies were worse than a random model in some situations.

*B. Q3 – There is an improvement between the LLaMA models from the first version to the second?*

Looking at the metrics of the models based on LLaMA 1, none managed to beat LLaMA 2's 13 billion parameters. The best LLaMA 1-based models achieved an accuracy of 40% for Portuguese (alpaca 13b) and 39.9% for English (Vicuna 7b), while LLaMA 2 13b achieved 46.8% for Portuguese and 49.3% for English. This was due to improvements in the base model, as described in [17]. This shows the capacity of open-source language models, and that they can improve even more overtime.

TABLE III
ACCURACY OF THE MODELS BY KNOWLEDGE AREA.

| Model | Humanities and its technologies | | Nature Sciences and its technologies | | Mathematics and its technologies | | Languages, codes and their technologies | |
|---|---|---|---|---|---|---|---|---|
| | Portuguese | English | Portuguese | English | Portuguese | English | Portuguese | English |
| **LLaMA 1 7b** | 0.221 | 0.292 | 0.227 | 0.209 | 0.217 | **0.262** | 0.233 | 0.262 |
| **LLaMA 1 13b** | 0.204 | 0.248 | 0.205 | 0.222 | 0.141 | 0.179 | 0.250 | 0.243 |
| **Alpaca 7b** | 0.233 | 0.201 | 0.205 | 0.213 | 0.128 | 0.141 | 0.208 | 0.240 |
| **Alpaca 13b** | 0.473 | 0.426 | 0.375 | 0.366 | 0.205 | 0.121 | 0.441 | 0.335 |
| **Koala 7B** | 0.180 | 0.201 | 0.196 | 0.196 | 0.121 | 0.134 | 0.212 | 0.215 |
| **Koala 13b** | 0.266 | 0.360 | 0.213 | 0.253 | 0.128 | 0.141 | 0.303 | 0.314 |
| **Vicuna 7b** | 0.384 | 0.508 | 0.262 | 0.331 | **0.243** | 0.198 | 0.356 | 0.434 |
| **Vicuna 13b** | 0.408 | 0.500 | 0.275 | 0.349 | **0.243** | 0.256 | 0.353 | 0.392 |
| **LLaMA 2 7b** | 0.304 | 0.381 | 0.270 | 0.340 | 0.211 | 0.179 | 0.211 | 0.339 |
| **LLaMA 2 13b** | **0.615** | **0.636** | **0.414** | **0.410** | 0.237 | **0.262** | **0.462** | **0.515** |

TABLE IV
MEAN AVERAGE INFERENCE TIME FOR THE MODELS (SECONDS).

| Model | AMD Ryzen 5 3600x | Intel i9 i9900k |
|---|---|---|
| **LLaMA 1 7b** | 18.3 | 16.8 |
| **LLaMA 1 13b** | 35.5 | 27.9 |
| **Alpaca 7b** | 20.5 | 17.2 |
| **Alpaca 13b** | 45.3 | 34.2 |
| **Koala 7B** | 21.4 | 15.9 |
| **Koala 13b** | 41.2 | 33.7 |
| **Vicuna 7b** | 21.7 | 17.0 |
| **Vicuna 13b** | 64.2 | 46.0 |
| **LLaMA 2 7b** | 17.5 | 13.7 |
| **LLaMA 2 13b** | 34.3 | 27.0 |

*C. Q4 – How efficient are the models in terms of time to run?*

Another factor of great importance in evaluating these models is the execution time of the inferences performed. To answer **Q4**, two experiments were conducted. In each of them, all models performed an inference for each of the questions in the data set. During the run, the times for performing the inferences (in seconds) were computed. Two machines were used, one equipped with an AMD Ryzen 5 3600x and the other equipped with an Intel i9 9900k. Table IV has the average times for running the questions.

Models with 13 billion parameters consistently take longer than models with 7 billion parameters. However, since these models do not require dedicated GPUs, these execution times are not prohibitive and allow the use of these LLMs by any interested party.

## VI. CONCLUSIONS AND FUTURE WORKS

This study presented a database for evaluating language models in Portuguese, offering a contribution to future research. In addition, we performed an evaluation of quantized language models that can be run on domestic hardware, expanding the dissemination and accessibility of these models, which represent a revolution in the field of natural language processing.

While the results may seem underwhelming, it is important to note that these language models are significantly smaller and have been trained with a smaller amount of data compared to the commercially available, closed-source options. Despite these limitations, the results indicate that the open-source models are progressing rapidly and are expected to improve their performance on tasks of this nature.

This paper is intended to provide a basis for future research, and therefore we present some ideas that emerged during the development of the study. They are:

- **Database expansion**: In order to restrict the scope of this study, only ENEM exams from the years 2010 to 2022 were considered. However, we believe that the generated scripts can be generalized to other years of ENEM, further expanding this database.
- **Evaluation of these models in other databases**: A similar task would be to evaluate these models on questions from public competitions. However, as there are many public exams each year, the exams from these competitions can be used to build an even more comprehensive and robust database.
- **Training Models**: The database provided contains a considerable amount of questions. It would be interesting to explore the possibility of training these language models to perform the task of answering questions.
- **Consider other models**: As shown in [9], there are already models trained for the purpose of explaining what their reasoning is for answering questions. Given this, future experiments can look in more depth at the rationale that led the model to a particular answer.
- **Consider multimodal models**: As shown in [12], the GPT-4 model performed impressively well on ENEM questions, in part due to its ability to process visual information in conjunction with the text. It is believed that multimodal models of this type will be available in open source in the near future.
- **Investigate the biases of the models**: Through the experiments conducted in this study, it was not possible to understand the reason for the observed biases in the behavior of the models. Therefore, in future investigations, this phenomenon can be further investigated.